\documentclass[11pt,a4paper]{article}
\usepackage[hyperref]{acl2021}
\usepackage{times}
\usepackage{latexsym}
\usepackage{amsmath}
\usepackage{booktabs}
\usepackage{enumitem}
\usepackage{multirow}
\usepackage{subcaption}

\newcommand\todo[1]{\colorbox{red!30}{TO-DO}}
\newcommand\expand[1]{{\colorbox{yellow!30}{Expand}}}

\usepackage{graphicx}
\usepackage{pifont}

\newcommand{\basep}{\textsc{baseline}}
\newcommand{\justp}{\textsc{justification}}
\newcommand{\justps}{\textsc{justification}}
\newcommand{\crowdp}{\textsc{crowd}}
\newcommand{\expertp}{\textsc{expert}}

\usepackage{tikz}
\makeatletter
\newcommand*{\radiobutton}{%
  \@ifstar{\@radiobutton0}{\@radiobutton1}%
}
\newcommand*{\@radiobutton}[1]{%
  \begin{tikzpicture}
    \pgfmathsetlengthmacro\radius{height("X")/2}
    \draw[radius=\radius] circle;
    \ifcase#1 \fill[radius=.6*\radius] circle;\fi
  \end{tikzpicture}%
}
\makeatother

\usepackage{microtype}

\aclfinalcopy 
\def\aclpaperid{2558} 

\title{What Ingredients Make for an Effective Crowdsourcing Protocol for Difficult NLU Data Collection Tasks?}

\author{Nikita Nangia$^1$\thanks{\ \ Equal contribution.}\And
        ~Saku Sugawara$^{2*}$\And
        Harsh Trivedi$^3$\\\AND\bf
        Alex Warstadt$^1$\And
        Clara Vania$^4$\thanks{\ \ Work done while at New York University.}\And
        Samuel R. Bowman$^1$ \\\AND
\textnormal{$^1$New York University,}
\textnormal{$^2$National Institute of Informatics,}
\textnormal{$^3$Stony Brook University,}
\textnormal{$^4$Amazon}\AND Correspondence: {\tt \{\href{mailto:nikitanangia@nyu.edu}{nikitanangia}, \href{mailto:bowman@nyu.edu}{bowman}\}@nyu.edu, \href{saku@nii.ac.jp}{saku@nii.ac.jp}}}

\date{}

\begin{document}
\maketitle
\begin{abstract}
Crowdsourcing is widely used to create data for common natural language understanding tasks. Despite the importance of these datasets for measuring and refining model understanding of language, there has been little focus on the crowdsourcing methods used for collecting the datasets.
In this paper, we compare the efficacy of interventions that have been proposed in prior work as ways of improving data quality.
We use multiple-choice question answering as a testbed and run a randomized trial by assigning crowdworkers to write questions under one of four different data collection protocols.
We find that asking workers to write explanations for their examples is an ineffective stand-alone strategy for boosting NLU example difficulty.
However, we find that training crowdworkers, and then using an iterative process of collecting data, sending feedback, and qualifying workers based on expert judgments is an effective means of collecting challenging data. But using crowdsourced, instead of expert judgments, to qualify workers and send feedback does not prove to be effective. 
We observe that the data from the iterative protocol with expert assessments is more challenging by several measures. Notably, the human--model gap on the unanimous agreement portion of this data is, on average, twice as large as the gap for the baseline protocol data.
\end{abstract}


\section{Introduction} 

\begin{figure*}[th]
\includegraphics[width=\linewidth]{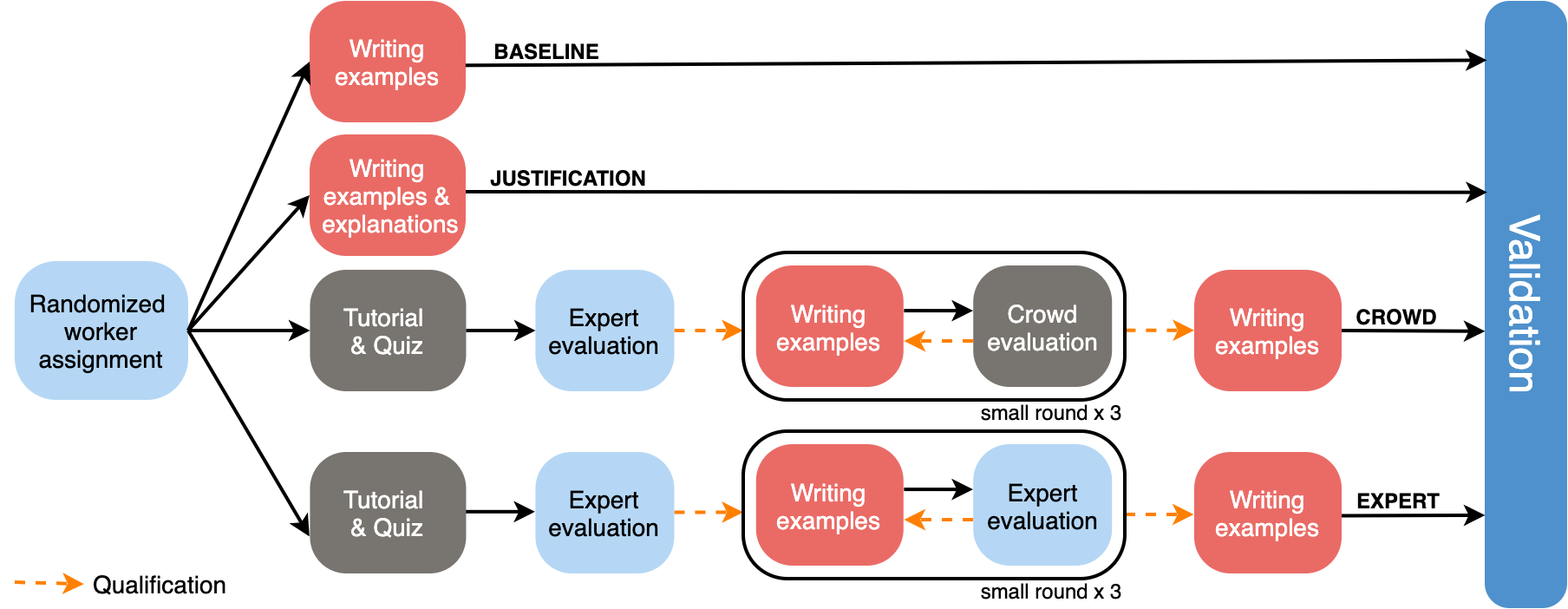}
\caption{The initial pool of crowdworkers are randomly assigned to one of four protocols and the datasets are collected in parallel.}
\label{fig:protocols}
\end{figure*}

Crowdsourcing is a scalable method for constructing examples for many natural language processing tasks.
Platforms like Amazon's Mechanical Turk give researchers access to a large, diverse pool of people to employ \cite{howe2006rise,snow-etal-2008-cheap,callison-burch-2009-fast}.
Given the ease of data collection with crowdsourcing, it has been frequently used for collecting datasets for natural language understanding (NLU) tasks like question answering \cite{mihaylov-etal-2018-suit}, reading comprehension \cite{rajpurkar-etal-2016-squad,huang-etal-2019-cosmos}, natural language inference \cite{dagan2005pascal,bowman-etal-2015-large,williams-etal-2018-broad,nie-etal-2020-adversarial}, and commonsense reasoning \cite{talmor-etal-2019-commonsenseqa}.

There has been substantial research devoted to studying crowdsourcing methods, especially in the human-computer interaction literature \cite{kittur-etal-2008-crowdsourcing, kittur-2011-crowdforge, bernstein-2012-analytic}.
However, most prior research investigates methods for collecting accurate \emph{annotations} for existing data, for example labeling objects in images or labeling the sentiment of sentences \cite{hsueh-etal-2009-data, liu-2019-interactive, sun-etal-2020-improving}.
There are some small-scale studies that use writing tasks, like writing product reviews, to compare crowdsourcing methodologies \citep{dow-2012-shepherd}. However, we are unaware of any prior work that directly evaluates the effects of crowdsourcing protocol design choices on the quality of the resulting data for NLU tasks.

Decisions around methodology and task design used to collect datasets dictate the quality of the data collected. As models become stronger and are able to solve existing NLU datasets, we have an increasing need for difficult, high-quality datasets that are still reliably solvable by humans. 
As a result, our thresholds for what makes a dataset acceptable become stricter: The data needs to be challenging, have high human-agreement, and avoid serious annotation artifacts \citep{gururangan-etal-2018-annotation}. To make collecting such large-scale datasets feasible, making well-informed crowdsourcing design decisions becomes crucial.

Existing NLP datasets have been crowdsourced with varying methods. The prevailing standard is to experiment with task design during pilots that are run before the main data collection \cite{vaughan2018making}.
This piloting process is essential to designing good crowdsourcing tasks with clear instructions, but the findings from these pilots are rarely discussed in published corpus papers, and the pilots are usually not large enough or systematic enough to yield definitive conclusions. 
In this paper, we use a randomized trial to directly compare crowdsourcing methodologies to establish general best practices for NLU data collection.

We compare the efficacy of three types of crowdsourcing interventions that have been used in previous work. 
We use multiple-choice question answering in English as a testbed for our study and collect four small datasets in parallel including a baseline dataset with no interventions. 
We choose QA as our test-bed over the similarly popular testbed task of natural language inference (NLI) because of our focus on very high human-agreement examples which calls for minimizing label ambiguity. In multiple-choice QA, the correct label is the answer choice that is \emph{most likely to be correct}, even if there is some ambiguity in whether that choice is genuinely true . In NLI however, if more than one label is plausible, then resolving the disagreement by ranking labels may not be possible \citep{pavlick-kwiatkowski-2019-inherent}.
In the trial, crowdworkers are randomly assigned to one of four protocols: \basep, \justps, \crowdp, or \expertp.\footnote{All the data is available at \href{https://github.com/nyu-mll/crowdsourcing-protocol-comparison}{https://github.com/nyu-mll/crowdsourcing-protocol-comparison}.} 
In \basep, crowdworkers are simply asked to write question-answering examples. In \justps they are tasked with also writing explanations for their examples, prompting self-assessment. For the \expertp\ and \crowdp\ protocols, we train workers using an iterative process of collecting data, sending feedback, and qualifying high performing workers to subsequent rounds. We use expert-curated evaluations in \expertp, and crowdsourced evaluations in \crowdp\ for generating feedback and assigning qualifications.
We use a a standard of high pay and strict qualifications for all protocols. We also validate the data to discard ambiguous and unanswerable examples. The experimental pipeline is sketched in Figure \ref{fig:protocols}. 

To quantify the dataset difficulty, we collect additional label annotations to establish human performance on each dataset and compare these to model performance. We also evaluate the difficulty of the datasets for typical machine learning models using IRT \cite{Baker1993ItemRT,lalor-etal-2016-building}. 

We find that the \expertp\ protocol dataset is the most challenging. The human--model gap with RoBERTa$_{\textsc{large}}$ \citep{liu2019roberta} on the unanimous agreement portion of \expertp\ is 13.9 percentage point, compared to 7.0 on the \basep\ protocol. The gap with UnifiedQA \citep{khashabi-etal-2020-unifiedqa} is 6.7 on \expertp, compared to 2.9 on \basep.
However, the \crowdp\ evaluation data is far less challenging than \expertp, suggesting that expert evaluations are more reliable than crowdsourced evaluations for sending feedback and assigning qualifications.

We also find that the \justps\ intervention is ineffective as a stand-alone method for increasing NLU data quality. A substantial proportion of the explanations submitted are duplicates, reused for multiple examples, or give trivial reasoning that is not specific to the example.

Lastly, to evaluate the datasets for serious annotation artifacts we test the guessability of answers by omitting the questions from the model input. This partial-input baseline achieves the lowest accuracy on \expertp, showing that the interventions used to successfully boost example difficulty may also reduce annotation artifacts.


\section{Related Work}
 
\paragraph{Creating NLU Corpora}
Existing NLU datasets have been collected using a multitude of methods, ranging from expert-designed, to crowdsourced, to automatically scraped.
The widely used Winograd schema dataset by \citet{levesque2012winograd} is constructed manually by specialists and it has 273 examples. 
Larger NLU datasets, more appropriate for training neural networks, are often crowdsourced, though the crowdsourcing methods used vary widely. 
Popular datasets, such as SQuAD \citep{rajpurkar-etal-2016-squad} for question answering and SNLI \citep{bowman-etal-2015-large} for natural language inference, are collected by providing crowdworkers with a context passage and instructing workers to write an example given the context.
\citet{rogers-etal-2020-getting} crowdsource QuAIL, a QA dataset, by using a more constrained data collection protocol where they require workers to write nine specific types of question for each passage.
QuAC \citep{choi-etal-2018-quac} is crowdsourced by pairing crowdworkers, providing one worker with a Wikipedia article, and instructing the second worker to ask questions about the hidden article. 

Recently, there has been a flurry of corpora collected using adversarial models in the crowdsourcing pipeline. \citet{dua-etal-2019-drop}, \citet{nie-etal-2020-adversarial}, and \citet{bartolo-etal-2020-beat} use models in the loop during data collection, where crowdworkers can only submit examples that cannot be solved by the models. However, such datasets can be biased towards quirks of the model used during data collection \cite{zellers-etal-2019-hellaswag, gardner-etal-2020-evaluating}.
%


\paragraph{Crowdsourcing Methods}
While crowdsourcing makes it easy to collect large datasets quickly, there are some clear pitfalls: Crowdworkers are generally less knowledgeable than field experts about the requirements the data needs to meet, crowdwork can be monotonous resulting in repetitive and noisy data, and crowdsourcing platforms can create a ``market for lemons" where fast work is incentivized over careful, creative work because of poor quality requesters \citep{akerlog-1978-market, chandler-2013-risks}. 

\citet{daniel2018quality} give a broad overview of the variables at play when trying to crowdsource high-quality data, discussing many strategies available to requesters. 
Motivated by the use of self-assessment in teaching \citet{boud-1995}, \citet{dow-2012-shepherd} study the effectiveness of self-assessment and external assessment when collecting data for product reviews. They find that both strategies are effective for improving the quality of submitted work. However, \citet{gadiraju2017using} find that crowdworker self-assessment can be unreliable since poor-performing workers overestimate their ability. \citet{drapeau-2016-microtalk} test a justify-reconsider strategy: Crowdworkers justify their annotations in a relation extraction task, they are shown a justification written by a different crowdworker, or an expert, and are asked to reconsider their annotation. They find that this method significantly boosts the accuracy of annotations.

Another commonly used strategy when crowdsourcing NLP datasets is to only qualify workers who pass an initial quiz or perform well in preliminary crowdsourcing batches  \citep{Wang2013PerspectivesOC, cotterell-callison-burch-2014-multi, ning-etal-2020-torque, shapira2020evaluating, roit-etal-2020-controlled}.
In addition to using careful qualifications, \citet{roit-etal-2020-controlled} send workers feedback detailing errors they made in their QA-SRL annotation. Writing such feedback is labor-intensive and can become untenable as the number of workers grows. \citet{dow-2011-managing} design a framework of promoting crowdworkers into ``shepherding roles" to crowdsource such feedback. We compare expert and crowdsourced feedback in our \expertp\ and \crowdp\ protocols.


\section{Data Collection Protocols}\label{sec:protocols}
We run our study on Amazon Mechanical Turk.\footnote{\url{https://www.mturk.com/}} At launch, crowdworkers are randomly assigned to one of four data collection protocols, illustrated in Figure \ref{fig:protocols}.\footnote{Screenshots of the task interfaces, and code to replicate them, are provided in the \href{https://github.com/nyu-mll/crowdsourcing-protocol-comparison}{git repository}.} 
To be included in the initial pool, workers need to have an approval rating of 98\% or higher, have at least 1,000 approved tasks, and be located in the US, the UK, or Canada.

\subsection{Writing Examples}\label{sec:baseline}
This task is used for collecting question-answer pairs in the crowdsourcing pipeline for all four protocols. Crowdworkers assigned to the \basep\ protocol are presented with only this task.

In this writing task, we provide a context passage drawn from the Open American National Corpus \cite{ide-suderman-2006-integrating}.\footnote{Following MultiNLI \cite{williams-etal-2018-broad}, we select the ten genres from OANC that are accessible to non-experts: Face-to-face, telephone, 911, travel, letters, slate, verbatim, government, OUP, and fiction.} Inspired by \citet{hu-etal-2020-ocnli}, we ask workers to write two questions per passage with four answer choices each. 
We direct workers to ensure that the questions are answerable given the passage and that there is only one correct answer for each question. We instruct them to limit word overlap between their answer choices and the passage and to write distracting answer choices that will seem plausibly correct to someone who hasn't carefully read the passage. To clarify these criteria, we provide examples of good and bad questions.

\subsection{Self-Assessment}\label{sec:self-just-inter}
Workers assigned to the \justps\ protocol are given the writing task described above (Section \ref{sec:baseline}) and are also tasked with writing a 1--3 sentence explanation for each question. They are asked to explain the reasoning needed to select the correct answer choice, mentioning what they think makes the question they wrote challenging.

\subsection{Iterative Feedback and Qualification}\label{sec:feedback-inter}

\paragraph{Tutorial}
Workers assigned to the \crowdp\ and \expertp\ protocols are directed to a tutorial upon assignment. The tutorial consists of two quizzes and writing tasks. 
The quizzes have four steps. In each step workers are shown a passage, two question candidates and are asked to select which candidate  (i) is less ambiguous, (ii) is more difficult, (iii) is more creative, or (iv) has better distracting answer choices.
These concepts are informally described in the writing task instructions, but the tutorial makes the rubric explicit, giving crowdworkers a clearer understanding of our desiderata.
We give workers immediate feedback on their performance during the first quiz and not the second so that we can use it for evaluation.
Lastly, for the tutorial writing tasks, we provide two passages and ask workers to write two questions (with answer choices) for each passage. These questions are graded by three experts\footnote{The expert annotators are authors of this paper and Dhara Mungra. All have research experience in NLU.} using a rubric with the same metrics described in the quiz, shown in Figure \ref{fig:rubric}. We give the qualification to continue onto the writing tasks to the top 60\% of crowdworkers who complete the tutorial. We only qualify the workers who wrote answerable, unambiguous questions, and we qualify enough workers to ensure that we would have a large pool of people in our final writing round.

\begin{figure}[t]
    \centering\small
    \fbox{%
        \parbox{0.95\linewidth}{
            \begin{minipage}{.95\linewidth}
            \smallskip
            \begin{enumerate}
                \item \textit{Is the question answerable and unambiguous?\smallskip} \\ \radiobutton\ Yes \ \radiobutton\ No \ \radiobutton\ Yes, but the label is wrong
                
                \item \textit{How closely do you think someone would need to read the passage to correctly answer the question?\smallskip} \\
                \radiobutton\ Wouldn't need to read it\\
                \radiobutton\ Quickly skim a few words or one sentence\\
                \radiobutton\ Quickly skim a few sentences\\
                \radiobutton\ Read the whole passage\\ 
                \radiobutton\ May need to read the passage more than once
                
                \item \textit{How creative do you think the question is?\smallskip}\\
                \radiobutton\ Not creative \ \radiobutton\ A little creative\\ \radiobutton\ Fairly creative \radiobutton\ Very creative
                
                \item \textit{Does the example have distracting answer choices?\smallskip}\\
                \radiobutton\ Yes \ \radiobutton\ No
            \medskip
            \end{enumerate}
        \end{minipage}
        }
    }
    \caption{The grading rubric used to evaluate examples submitted during the intermediate writing rounds in the \expertp\ and \crowdp\ protocols.}
    \label{fig:rubric}
\end{figure}

\paragraph{Intermediate Writing Rounds}
After passing the tutorial, workers go through three small rounds of writing tasks. At the end of each round, we send them feedback and qualify a smaller pool of workers for the next round. We only collect 400--500 examples in these intermediate rounds. At the end of each round, we evaluate the submitted work using the same rubric defined in the tutorial. In the \expertp\ protocol, three experts grade worker submissions, evaluating at least four questions per worker. 
The evaluation annotations are averaged and workers are \textbf{qualified} for the next round based on their performance. The qualifying workers are sent a message with \textbf{feedback} on their performance and a bonus for qualifying. Appendix \ref{app:feedback} gives details on the feedback sent.

Evaluating the examples in each round is labor-intensive and challenging to scale (avg. 30 expert-min. per worker). In the \crowdp\ protocol we experiment with crowdsourcing these evaluations. After the first intermediate writing round in \crowdp, experts evaluate the submitted work. The evaluations are used to qualify workers for the second writing round \emph{and} to promote the top 20\% of workers into a feedback role.
After intermediate writing rounds 2 and 3, the promoted workers are tasked with evaluating all the examples (no one evaluates their own work). We collect five evaluations per example and use the averaged scores to send feedback and qualify workers for the subsequent round.

For both \crowdp\ and \expertp\ protocols, the top 80\% of workers are requalified at the end of each round. Of the 150 workers who complete the tutorial, 20\% qualify for the final writing round. Our qualification rate is partly dictated by a desire to have a large enough pool of people in the final writing task to ensure that no dataset is skewed by only a few people \citep{geva-etal-2019-modeling}.

\paragraph{Cost} 
We aim to ensure that our pay rate is at least US \$15/hr for all tasks. 
The total cost per question, excluding platform fees, is \$1.75 for the \basep\ protocol and \$2 for \justps. If we discard all the data collected in the intermediate writing rounds, the cost is \$3.76 per question for \expertp,\footnote{The discarded data collected during training was annotated by experts, and if we account for the cost of expert time used, the cost for \expertp\ increases to \$4.23/question. This estimate is based on the approximate hourly cost of paying a US PhD student, including benefits and tuition.} and \$5 for \crowdp.

The average pay given during training to workers that qualify for the final writing task in \expertp\ is about \$120/worker (with an estimated 6--7 hours spent in training). In \crowdp, there is an additional cost of \$85/worker for collecting crowdsourced evaluations. The cost per example, after training, is \$1.75 per question for both protocols, and total training cost does not scale linearly with dataset size, as one may not need twice as many writers for double the dataset size. More details on our payment and incentive structure can be found in Appendix \ref{app:payment}.


\begin{table*}[ht]
\centering
\small
\begin{tabular}{ l c c r|l r|l }
\toprule
    Dataset & \emph{N} & Human & RoBERTa & $\Delta$ & UniQA & $\Delta$ \\
    \midrule
    
	\basep\ & \emph{1492} & - & 88.8 (0.2) & - & 93.6 & -  \\
	\justp\ & \emph{1437} & - & 86.5 (0.6) & - & 91.4 & -  \\
	\crowdp\ & \emph{1544} & - & 81.8 (0.7) & - & 88.1 & -  \\
	\expertp\ & \emph{1500} & - & 81.3 (0.6) & - & 87.7 & -  \\
    
    \midrule[.06em]\midrule[.04em]
    \emph{Results on the 10-way annotated subset} \\
    \midrule[.04em]\midrule[.06em]
    
	\basep\ & \emph{482} & 95.9 & 87.2 (0.8) & \hphantom{0}8.7 & 92.5 & 3.3  \\
	\justp\ & \emph{471} & 95.5 & 86.7 (1.0) & \hphantom{0}8.9 &  90.9 & \textbf{4.7} \\
	\crowdp\ & \emph{472} & 94.8 & 83.5 (1.0) & 11.3 & 90.5 & \textbf{4.3}  \\
	\expertp\ & \emph{464} & 92.8 & 80.6 (1.1) & \textbf{12.2} & 89.8 & 3.0 \\

    \midrule[.04em]
    \emph{High agreement ($>$80\%) portion of 10-way annotated data} \\
    \midrule[.04em]

	\basep\ & \emph{436} & 97.7 & 89.3 (0.8) & \hphantom{0}8.4 & 94.0 & 3.7  \\
	\justp\ & \emph{419} & 97.8 & 89.5 (0.6) & \hphantom{0}8.3 & 93.1 & \textbf{4.8} \\
	\crowdp\ & \emph{410} & 96.8 & 86.2 (0.9) & 10.6 & 93.6 & 3.2  \\
	\expertp\ & \emph{383} & 98.2 & 84.7 (1.3) & \textbf{13.5} & 92.9 & \textbf{5.3}  \\
    
    \midrule[.04em]
    \emph{Unanimous agreement portion of 10-way annotated data} \\
    \midrule[.04em]

	\basep\ & \emph{340} & 99.1 & 92.1 (0.7) & \hphantom{0}7.0 & 96.2 & 2.9  \\
	\justp\ & \emph{307} & 98.7 & 93.2 (0.3) & \hphantom{0}5.5 & 95.8 & 2.9  \\
	\crowdp\ & \emph{277} & 98.6 & 88.9 (0.9) & \hphantom{0}9.7 & 97.1 & 1.4 \\
	\expertp\ & \emph{271} & 99.3 & 85.4 (1.1) & \textbf{13.9} &  92.5 & \textbf{6.7}  \\

\bottomrule
\end{tabular}
\caption{Human and model performance on each of our datatsets. \emph{N} shows the number of examples in the dataset. 
\emph{RoBERTa} shows average zero-shot performance for six RoBERTa$_{\textsc{large}}$ models finetuned on RACE, standard deviation is in parentheses. \emph{UniQA} shows zero-shot performance of the T5-based UnifiedQA-v2 model.
\emph{$\Delta$} shows the differences in human and model performance.}
\label{tbl:human-model-gap}
\end{table*}


\section{Data Validation}
We collect label annotations by asking crowdworkers to pick the correct answer choice for a question, given the context passage. In addition to the answer choices written by the writer, we add an \emph{Invalid question / No answer} option.
We validate the data from each protocol. For \crowdp\ and \expertp, we only validate the data from the final large writing rounds. Data from all four protocols is shuffled and we run a single validation task, collecting either two or ten annotations per example. 

We use the same minimum qualifications as the writing task (Section \ref{sec:protocols}), and require that workers first pass a qualification task. The qualification task consists of 5 multiple-choice QA examples that have been annotated by experts.\footnote{These examples are taken from intermediate rounds 1, 2, and 3 of the \expertp\ protocol.} People who answer at least 3 out of 5 questions correctly receive the qualification to work on the validation tasks. Of the 200 crowdworkers who complete the qualification task, 60\% qualify for the main validation task. Following \citet{ho2015incentivizing}, to incentivize higher quality annotations, we include expert labeled examples in the validation task, constituting 10\% of all examples. If a worker's annotation accuracy on these labeled examples falls below 50\%, we remove their qualification (7 workers are disqualified through this process), conversely workers who label these examples correctly receive a bonus.

\paragraph{10-Way Validation} \citet{pavlick-kwiatkowski-2019-inherent} show that annotation disagreement may not be noise, but could be a signal of true ambiguity. \citet{nie-etal-2020-learn} recommend using high-human-agreement data for model evaluation to avoid such ambiguity. To have enough annotations to filter the data for high human agreement and to estimate human performance, we collect ten annotations for 500 randomly sampled examples per protocol.

\paragraph{Cost}
We pay \$2.50 for the qualification task and \$0.75 per pair of questions for the main validation task. For every 3 out of 4 expert-labeled examples a worker annotates correctly, we send a \$0.50 bonus. 

\section{Datasets and Analysis}

We collect around 1,500 question-answer pairs from each protocol design: 1,558 for \basep, 1,534 for \justps, 1,600 for \crowdp, and 1,580 for \expertp. We use the validation annotations to determine the gold-labels and to filter out examples: 
If there is no majority agreement on the answer choice, or if the majority selects \emph{invalid question}, the example is discarded ($\sim5\%$ of examples). 
For the 2-way annotated data, we take a majority vote over the two annotations plus the original writer's label. For the 10-way annotated data, we sample four annotations and take a majority vote over those four plus the writer's vote, reserving the remainder to compute an independent estimate of human performance.

\subsection{Human Performance and Agreement}\label{sec:human-perf}
For the 10-way annotated subsets of the data, we take a majority vote over the six annotations that are \textit{not} used when determining the gold answer, and compare the result to the gold answer to estimate human performance. 
Table \ref{tbl:human-model-gap} shows the result for each dataset. 
The \expertp\ and \crowdp\ datasets have lower human performance numbers than \basep\ and \justps. This is also mirrored in the inter-annotator agreement for validation, where Krippendorf's~$\alpha$ \cite{krippendorff1980content} is 0.67 and 0.71 for \expertp\ and \crowdp, compared to 0.81 and 0.77 for \basep\ and \justps\ (Table \ref{tbl:agreement} in Appendix \ref{app:agreement}).
The lower agreement may be reflective of the fact that while these examples are still clearly human solvable, they are more challenging than those in \basep\ and \justps\ As a result, annotators are prone to higher error rates, motivating us to look at the higher agreement portions of the data to determine true dataset difficulty. 
And while the agreement rate is lower for \expertp\ and \crowdp, more than 80\% of the data still has high human-agreement on the gold-label, where at least 4 out of 5 annotators agree on the label. 
The remaining low-agreement examples may have more ambiguous questions, and we follow \citeauthor{nie-etal-2020-learn}'s (\citeyear{nie-etal-2020-learn}) recommendation and focus our analysis on the high-agreement portions of the dataset.

\subsection{Zero-Shot Model Performance}
We test two pretrained models that perform well on other comparable QA datasets: RoBERTa$_{\textsc{large}}$ \cite{liu2019roberta} and UnifiedQA-v2 \cite{khashabi-etal-2020-unifiedqa}.
We fine-tune RoBERTa$_{\textsc{large}}$ on RACE \cite{lai-etal-2017-race}, a large-scale multiple-choice QA dataset that is commonly used for training  \cite{sun-etal-2019-improving}. We fine-tune 6 RoBERTa$_{\textsc{large}}$ models and report the average performance across runs. The UnifiedQA-v2 model is a single T5-based model that has been trained on 15 QA datasets.\footnote{The authors of UnifiedQA kindly shared the unreleased v2 model with us.}
We also fine-tune RoBERTa$_{\textsc{large}}$ on CosmosQA and QuAIL, finding that zero-shot model performance is best with RACE fine-tuning but that the trends in model accuracy across our four datasets are consistent (Appendix \ref{app:zero-shot}).

\subsection{Comparing Protocols} 
As shown in Table \ref{tbl:human-model-gap}, model accuracy on the full datasets is lowest for \expertp, followed by \crowdp, \justps, and then \basep. However, model accuracy alone does not tell us how much headroom is left in the datasets. Instead, we look at the difference between the estimated human performance and model performance.

\paragraph{Human--Model gap} The trends in the human--model gap on the 10-way annotated sample are inconsistent across models. For a more conclusive analysis, we focus on the higher-agreement portions of the data where label ambiguity is minimal.

On the high agreement section of the datasets, both models' performance is weakest on \expertp. RoBERTa$_{\textsc{large}}$ shows the second largest human--model gap on \crowdp, however for UnifiedQA \justps\ is the next hardest dataset. This discrepancy between the two types of iterative feedback protocols is even more apparent in the unanimous agreement portion of the data. On the unanimous agreement examples, both models show the lowest performance on \expertp\ but Unified-QA achieves near perfect performance on \crowdp. This suggests that while the \crowdp\ protocol used nearly the same crowdsourcing pipeline as \expertp, the evaluations done by experts are a much more reliable metric for selecting workers to qualify and for generating feedback, at the cost of greater difficulty with scaling to larger worker pools. This is confirmed by inter-annotator agreement: Expert agreement on the rubric-based evaluations has a Krippendorf's 
$\alpha$ of 0.65, while agreement between crowdworker evaluations is 0.33.

\paragraph{Self-Justification} Model performance on the unanimous agreement examples of \justps\ is comparable to, or better than, performance on \basep. 
To estimate the quality of justifications, we manually annotate a random sample of 100 justifications. 
About 48\% (95\% CI: [38\%, 58\%]) are duplicates or near-duplicates of other justifications, and of this group, nearly all are trivial (e.g. \emph{Good and deep knowledge is needed to answer this question}) and over half are in non-fluent English (e.g. \emph{To read the complete passage to understand the question to answer.}). On the other hand, non-duplicate justifications are generally of much higher quality, mentioning distractors, giving specific reasoning, and rewording phrases from the passage (e.g. \emph{Only \#1 is discussed in that last paragraph. The rest of the parts are from the book, not the essay. Also the answer is paraphrased from ``zero-sum" to ``one's gain is another's loss"}). 
While we find that \justps\ does not work as a stand-alone strategy, we cannot conclude that self-justification would be equally ineffective if combined with more aggressive screening to exclude crowdworkers who author trivial or duplicate justifications. \citet{gadiraju2017using} also recommend using the accuracy of a worker's self-assessments to screen workers.



\paragraph{Cross-Protocol Transfer}\label{sec:cross-protocol}
Since the datasets from some protocols are clearly more challenging than others, it prompts the question: are these datasets also better for training models? To test cross-protocol transfer, we fine-tune RoBERTa$_{\textsc{large}}$ on one dataset and evaluate on the other three. 
We find that model accuracy is not substantively better from fine-tuning on any one dataset (Table \ref{tbl:cross-validation}, Appendix \ref{app:cross-protocol}). The benefit of \expertp\ being a more challenging evaluation dataset does not clearly translate to training. 
However, these datasets may be too small to offer clear and distinguishable value in this setting.

\paragraph{Annotation Artifacts} 

\begin{table}[t!]
\small\centering
\begin{tabular}{lccc}\toprule
    Partial input  & P + A & Q + A & A \\ \midrule
    \basep\ & 69.9 (4.7) & 41.9 (2.9) & 34.9 (2.4) \\
    \justps\ & 57.9 (1.3) & 38.3 (2.2) & 33.9 (6.3) \\
    \crowdp\ & 57.7 (3.1) & 43.9 (2.0) & 35.2 (1.9) \\
    \expertp\ & 52.0 (1.5) & 42.8 (1.8) & 35.7 (1.4) \\
    \bottomrule
\end{tabular}
\caption{Accuracy (std.) of partial input baselines. \emph{P} is passage, \emph{Q} is question, and \emph{A} is answer choices.}
\label{tbl:partial-input-performance-race}
\end{table}

To test for undesirable artifacts, we evaluate partial input baselines \citep{kaushik-lipton-2018-much, poliak-etal-2018-hypothesis}. We take a RoBERTa$_{\textsc{large}}$ model, pretrained on RACE, and fine-tune it using five-fold cross-validation, providing only part of the example input. We evaluate three baselines: providing the model with the passage and answer choices only, the question and answer choices only, and the answer choices alone. Results are shown in Table \ref{tbl:partial-input-performance-race}. The passage+answer baseline has significantly lower performance on the \expertp\ dataset in comparison to the others. This indicates that the iterative feedback and qualification method using expert assessments not only increases overall example difficulty but may also lower the prevalence of simple artifacts that can reveal the answer. Performance of the question+answer and answer-only baselines is comparably low on all four datasets. 

\paragraph{Question and Answer Length}
\begin{figure}[t!]
\includegraphics[width=\linewidth]{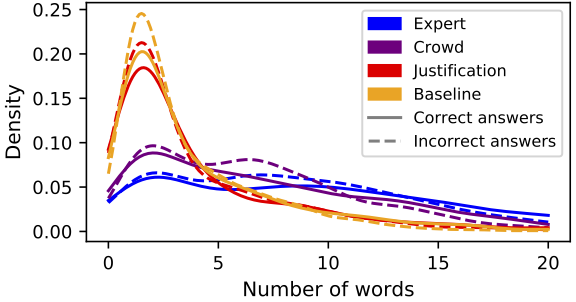}
\caption{Distribution of answer lengths. The distributions for different datasets and for the correct and incorrect answer options are plotted separately.}
\label{fig:answer-length}
\end{figure}

We observe that the difficulty of the datasets is correlated with average answer length (Figure \ref{fig:answer-length}). The hardest dataset, \expertp, also has the longest answer options with an average of 9.1 words, compared to 3.7 for \basep, 4.1 for \justps, and 6.9 for \crowdp. This reflects the tendency of the 1- and 2-word answers common in the \basep\ and \justps\ datasets to be extracted directly from the passage. While sentence-length answers, more common in \expertp\ and \crowdp, tend to be more abstractive.
Figure \ref{fig:answer-length} also shows that incorrect answer options tend to be shorter than correct ones. This pattern holds across all datasets, suggesting a weak surface cue that models could exploit. Using an answer-length based heuristic alone, accuracy is similar to the answer-only model baseline: 34.2\% for \basep, 31.7\% for \justps, 31.5\% for \crowdp, and 34.3\% for \expertp.

\paragraph{Wh-words} We find that the questions in \expertp\ and \crowdp\ protocols have similar distributions of wh-words, with many \emph{why} questions and few \emph{who} or \emph{when} questions compared to the \basep\ and \justps\ protocols, seemingly indicating that this additional feedback prompts workers to write more complex questions.

\paragraph{Non-Passage-Specific Questions}
We also observe that many questions in the datasets are formulaic and include no passage-specific content, for instance \emph{Which of the following is true?}, \emph{What is the main point of the passage?}, and \emph{Which of the following is not mentioned in the passage?}. We manually annotate 200 questions from each protocol for questions of this kind. We find that there is no clear association between the dataset's difficulty and the frequency of such questions: 15\% of questions in \expertp\ are generic, compared to 4\% for \crowdp, 10\% for \justps, and 3\% for \basep. We might expect that higher quality examples that require reading a passage closely would ask questions that are specific rather than generic. 
But our results suggest that difficulty may be due more to the subtlety of the answer options, and the presence of distracting options, rather than the complexity or originality of the questions.

\paragraph{Order of Questions}

We elicit two questions per passage in all four protocols with the hypothesis that the second question may be more difficult on aggregate. However, we find that there is only a slight drop in model accuracy from the first to second question on the \crowdp\ and \expertp\ datasets (1.0 and 0.7 percentage points). And model accuracy on \basep\ remains stable, while it increases by 2.7 percentage points on \justps. A task design with minimal constraints, like ours, does not prompt workers to write an easier question followed by a more difficult one, or vice versa.

\subsection{Item Response Theory} 

Individual examples within any dataset can have different levels of difficulty. To better understand the distribution of difficult examples in each protocol, we turn to Item Response Theory \citep[IRT;][]{Baker1993ItemRT}, which has been used to estimate individual example difficulty based on model responses \citep{lalor-etal-2019-learning,MARTINEZPLUMED201918}. Specifically, we use the three-parameter logistic (3PL) IRT model, where an example is characterized by discrimination, difficulty, and guessing parameters. Discrimination defines how effective an example is at distinguishing between weak and strong models, difficulty defines the minimum ability of a model needed to obtain high performance, and the guessing parameter defines the probability of a correct answer by random guessing. Following \citet{Vania21IRT}, we use 90 Transformer-based models fine-tuned on RACE, with varying ability levels, and use their predictions on our four datasets as responses. For comparison, we also use model predictions on QuAIL and CosmosQA. Refer to Appendix \ref{app:irt} for more details.

\begin{figure}[t!]
    \includegraphics[width=\linewidth]{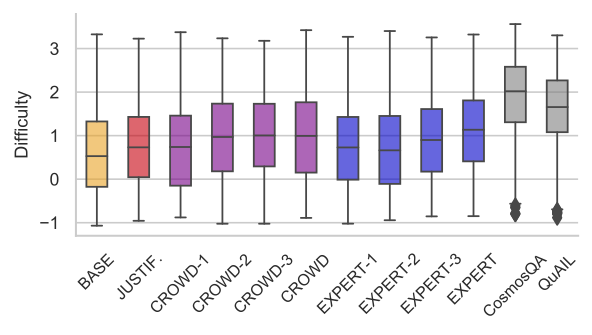}
    \caption{Distribution of examples according to their difficulty parameters. \textsc{crowd}/\textsc{expert}-$\{1,2,3\}$ are the three intermediate rounds of data that are not included in the final datasets.}
\label{fig:irt-difficulty}
\end{figure}

Figure \ref{fig:irt-difficulty} shows the distribution of example difficulty for each protocol. Also plotted are the difficulty parameters for the intermediate rounds of data that are collected in the iterative feedback protocols.\footnote{The IRT parameters for discrimination range from 0.6 to 2.1, while for guessing they range from 0.03 to 0.74. However, we observe that the distributions of both parameters across the four datasets are similar.} We see that \expertp\ examples have the highest median and 75th percentile difficulty scores, while \basep\ scores the lowest. We also note that the greatest gain in difficulty for \crowdp\ examples happens between rounds 1 and 2, the only feedback and qualification stage that is conducted by experts. This offers further evidence that expert assessments are more reliable, and that crowdsourcing such assessments poses a significant challenge.

While the examples in \expertp\ have higher difficulty scores than the other protocols, the scores are significantly lower than those for CosmosQA and QuAIL (all four datasets show similar discrimination scores to CosmosQA and QuAIL). The data collection methods used for both CosmosQA and QuAIL differ substantially from methods we tested. \citet{rogers-etal-2020-getting} constrain the task design for QuAIL and require workers to write questions of specific types, like those targeting temporal reasoning. Similarly, in CosmosQA workers are encouraged to write questions that require causal or deductive commonsense reasoning. In contrast, we avoid dictating question type in our instructions. The IRT results here suggest that using prior knowledge to slightly constrain the task design can be effective for boosting example difficulty. In addition to differing task design, CosmosQA and QuAIL also use qualitatively different sources for passages. Both datasets use blogs and personal stories, QuAIL also uses texts from published fiction and news. 
Exploring the effect of source text genre on crowdsourced data quality is left to future work.

\section{Conclusion}
We present a study to determine effective protocols for crowdsourcing difficult NLU data. We run a randomized trial to compare interventions in the crowdsourcing pipeline and task design. 
Our results suggest that asking workers to write justifications is not a helpful stand-alone strategy for improving NLU dataset difficulty, at least in the absence of explicit incentives for workers to write high-quality justifications. However, we find that training workers using an iterative feedback and requalification protocol is an effective strategy for collecting high-quality QA data. The benefit of this method is most evident in the high-agreement subset of the data where label noise is low. We find that using expert assessments to conduct this iterative protocol is fruitful, in contrast with crowdsourced assessments that have much lower inter-annotator agreement and the noisy signal from these assessments does not boost example difficulty.

\section*{Acknowledgements}
We thank Dhara Mungra for her early contributions to this project, and for being one of the expert graders during data collection. We also thank Daniel Khashabi for giving us access to UnifiedQA-v2 for our experiments.
This work has benefited from financial support to SB by Eric and Wendy Schmidt (made by recommendation of the Schmidt Futures program),  Apple, and Intuit, and from in-kind support by the NYU High-Performance Computing Center and by NVIDIA Corporation (with the donation of a Titan V GPU). 
SS was supported by JST PRESTO Grant No. JPMJPR20C4.
This  material  is  based  upon  work  supported  by the National Science Foundation under Grant No. 1922658. Any opinions, findings, and conclusions or recommendations expressed in this material are those of the author(s) and do not necessarily reflect the views of the National Science Foundation.

\section*{Ethics Statement}
We are cognizant of the asymmetrical relationship between requesters and workers in crowdsourcing, and we take care to be responsive employers and to pay a wage commensurate with the high-quality work we're looking for. So in additional to the ethical reasons for paying fair wages, our successes with collecting high-quality NLU data offer weak evidence that others should also follow this practice. However, the mere existence of more research on NLU crowdsourcing with positive results could arguably encourage more people to do crowdsourcing under a conventional model, with low pay and little worker recourse against employer malpractice.
The only personal information we collect from workers is their Mechanical Turk worker IDs, which we keep secure and will not release. However, we do not engage with issues of bias during data collection and we expect that the data collected under all our protocols will, at least indirectly, reinforce stereotypes. 

We confirmed with New York University's IRB that crowdsourced NLP dataset construction work, \emph{including} experimental work on data collection methods, is exempt from their oversight. The only personal information we collect from workers is their Mechanical Turk worker IDs, which we keep secure and will not release.

\bibliographystyle{acl_natbib}
\bibliography{anthology,acl2021}

\appendix
\clearpage

\section{Iterative Protocol Feedback}\label{app:feedback}
In the \expertp\ and \crowdp\ protocols, we conduct three small intermediate rounds of data collection to help train crowdworkers and give them feedback on their submissions. At the end of each small round of writing, the submitted examples are evaluated either by experts or crowdworkers, as described in Section \ref{sec:feedback-inter}. The rubric given in Figure \ref{fig:rubric} is used during evaluations. After compiling the evaluations, we qualify the top 80\% of workers for the next round and send them a feedback message. We tell workers what their difficulty and creativity scores are in comparison to the average. We also tell them what percentage of their question-answer pairs were labeled as having distracting answer choices and what percentage were labeled ambiguous, with examples of any such questions. Lastly, we list the examples they wrote that received the highest and lowest overall rubric scores.

\section{Payment and Incentive Structure}\label{app:payment}
The compensation for for writing two questions in the baseline writing task is \$3.50, excluding platform fees, we estimate it takes 12--15 minutes to do a close reading of the passage and write two challenging questions. For the \justps\ protocol, the compensation is \$4 per task to account for the additional time it takes to write a justifications for each question. 
For the tutorial that workers in the \crowdp\ and \expertp\ protocols need to complete, we pay \$3.50, and give a bonus of \$1.50 if they qualify onto the writing tasks. Similarly, at the end of each intermediate writing batch, a bonus is sent to the workers that qualify for the subsequent round: \$5, \$7, and \$10 after the 1st, 2nd and 3rd rounds respectively.
Promoted workers who are tasked with the crowdsourced evaluations in the \crowdp\ protocol, are paid \$0.50 per question. They are also sent a bonus of \$5 for each round of evaluations they complete.

\section{Inter-Annotator Agreement}\label{app:agreement}
Table \ref{tbl:agreement} shows the inter-annotator agreement during data validation task for each dataset. The Krippendorf's $\alpha$ is lowest for \expertp, which also has the lowest human performance baseline, likely due to the pressure to produce subtle questions.

\begin{table}[h!]
    \centering\small
    \begin{tabular}{lcccc} \toprule
    Protocol & $\alpha_{all}$ & $\alpha_{10}$ \\ \midrule
    \basep\  &  0.81 & 0.79 \\
    \justps\  & 0.77 & 0.74 \\
    \crowdp\  & 0.71 & 0.69 \\
    \expertp\ & 0.67 & 0.64 \\
    \bottomrule
    \end{tabular}
    \caption{Inter-annotator agreement statistics for each datatset.  $\alpha_{all}$ and $\alpha_{10}$ give the Krippendorf's $\alpha$ scores for all examples and the subset of 10-way annotated examples respectively.}
    \label{tbl:agreement}
\end{table}

\section{Zero-Shot Model Performance: CosmosQA and QuAIL}\label{app:zero-shot}
In addition to fine-tuning RoBERTa$_{\textsc{large}}$ on RACE, we also fine-tune it on CosmosQA, and QuAIL to test zero-shot model performance. Table \ref{tbl:cosmos-quail} shows the zero-shot results. We observe that model performance on our datasets is substantially worse when fine-tuning on CosmosQA or QuAIL. However, the pattern in model behaviour is consistent regardless of corpus used. In all three conditions, model accuracy is highest on \basep, followed by \justps, then \crowdp, and finally \expertp.

\begin{table}[t!]
    \centering\small
    \begin{tabular}{lccc} 
    \toprule
    Dataset & RACE & CosmosQA & QuAIL \\ \midrule
    \basep\  &  88.8 & 74.1 & 80.5 \\
    \justps\  & 86.5 & 65.9 & 68.8 \\
    \crowdp\  & 81.8 & 65.1 & 62.7 \\
    \expertp\ & 81.3 & 56.8 & 52.4 \\
    \bottomrule
    \end{tabular}
\caption{Zero-shot model accuracy on our datasets, when training on the datasets named in the columns.}
\label{tbl:cosmos-quail}
\end{table}

\section{Cross-Protocol Transfer}\label{app:cross-protocol}
As discussed in Section \ref{sec:cross-protocol}, we test cross-protocol transfer by fine-tuning RoBERTa$_{\textsc{large}}$ on one dataset and evaluating on the other three. For a baseline comparison, we also fine-tune the model on each dataset using five-fold cross-validation. Results are shown in Table \ref{tbl:cross-validation}.

\begin{table}[h!] 
    \centering \small
    \begin{tabular}{lcccc|c} \toprule
      & \textsc{base} & \textsc{just} & \textsc{crowd} & \textsc{exp} & Cross-val\\ \midrule
    
    
    \textsc{base} & - & 88.2 & 87.4 & 87.8 & 87.9 (2.0) \\
    \textsc{just} & 84.9 & - & 85.3 & 84.9 & 85.6 (2.4) \\
    \crowdp\ & 81.6 & 83.2 & - & 81.7 & 82.5 (1.9) \\
    \expertp\ & 80.6 & 81.2 & 81.7 & - & 82.8 (1.4) \\
    
    \bottomrule
    \end{tabular}
    \caption{Cross-protocol evaluation where the row and column indicate target and source datasets respectively. \emph{Cross-val} shows the accuracy and std. dev. from five-fold cross-validation on each dataset.}
    \label{tbl:cross-validation}
\end{table}

\section{IRT Setup}\label{app:irt}

\paragraph{IRT Model}
We use the 3PL IRT model, where the probability of a responder $i$ of answering an item $j$ is given as:

\begin{align*}
    p_j(\theta_i) = \gamma_j + \frac{1-\gamma_j}{1 + e^{-\alpha_j(\theta_i-\beta_j))}}
\end{align*}

where $\alpha, \beta, \gamma$ denote the discrimination, the difficulty, and the guessing parameters, respectively. Following \citet{lalor-etal-2019-learning}, we use variational inference (VI) to estimate these parameters. Given a set of model responses $M$, we use the following variational posterior to estimate the joint probability of the parameters $\pi(\theta, \alpha, \beta, \gamma \mid M)$:

\begin{align*}
  q(\theta, \alpha, \beta, \gamma) = \prod_{i=1}^I \pi_i^\theta(\theta_i) \prod_{j=1}^J \pi_j^\alpha(\alpha_i) \pi_j^\beta(\beta_i) \pi_j^\gamma(\gamma_i),
\end{align*}

where $\pi^\rho(\cdot)$ is the density for parameter $\rho$. We use the following distributions for each parameter:
$\mathcal{N}(\mu_\theta, \sigma_\theta^2)$ for $\theta$,
$\mathcal{N}(\mu_\alpha, \sigma_\alpha^2)$ for $\log \alpha$,
$\mathcal{N}(\mu_\beta, \sigma_\beta^2)$ for $\beta$, and 
$\mathcal{N}(\mu_\gamma, \sigma_\gamma^2)$ for $\mathrm{sigmoid}^{-1}(\gamma)$.
We then fit the posterior parameters by minimizing the KL divergence between $q(\theta, \alpha, \beta, \gamma)$ and the true posterior $\pi(\theta, \alpha, \beta, \gamma \mid Y)$. This is equivalent to minimizing the evidence lower bound (ELBO). 

To control for different test sizes, we weight the log likelihood of each item's parameter by the inverse of the item's test size when fitting the parameters. We adapt prior used by \citet{lalor-etal-2019-learning} for each parameter: $\mathcal{N}(0, 1)$ for $\theta$, $\beta$, and $\mathrm{sigmoid}^{-1}(\gamma)$. For $\log \alpha$, we use $\mathcal{N}(0, \sigma_\alpha^2)$ where we set $\sigma_\alpha$ by searching $[0.25, 0.5]$ by increments of $0.05$ and use the value yielding the highest ELBO.

\paragraph{Pretrained Transformer Models}

We use 18 Transformer-based models: ALBERT-XXL-v2~\citep{Lan2020ALBERT}, RoBERTa$_{\textsc{large}}$ and RoBERTa$_\textsc{base}$~\citep{liu2019roberta}, BERT$_{\textsc{large}}$ and BERT$_{\textsc{base}}$~\citep{devlin-etal-2019-bert}, XLM-R~\citep{conneau-etal-2020-unsupervised}, and 12 MiniBERTas~\citep{Zhang20MiniBERTas}.\footnote{We use pretrained models distributed with HuggingFace Transformers~\citep{wolf-etal-2020-transformers}.} We fine-tune each of these models on RACE, and keep five different checkpoints---at 1\%, 10\%, 25\%, and 50\% of the maximum training epochs, plus the best checkpoint on the RACE validation set. In total, we have 90 model responses for each test example. For all the models, we use a batch size of 8, learning rate of $1.0 \times 10^{-5}$, and finetune the models using the Adam optimizer for 4 epochs on the RACE dataset.



\end{document}